\documentclass[letterpaper, 10 pt, conference]{ieeeconf}  

\IEEEoverridecommandlockouts                              

\usepackage{graphicx}
\usepackage{caption}

\graphicspath{
    {figures/}
}

\usepackage{amsmath,amssymb,enumerate}

\usepackage{amsthm}
\usepackage{setspace}
\usepackage{booktabs}
\usepackage{algorithm}
\usepackage{algpseudocode}
\usepackage{amsmath}

\usepackage[usenames,dvipsnames,svgnames,table]{xcolor}
\usepackage{mathtools}
\usepackage{blindtext}
\usepackage{gensymb}
\usepackage{xparse}
\usepackage{lipsum}
\usepackage{mathrsfs}
\usepackage[mathscr]{euscript}
\usepackage{times}
\usepackage{cite} 
\usepackage{multicol}
\definecolor{darkgreen}{rgb}{0,0.6,0}
\usepackage[bookmarks=true,colorlinks=true,pdfpagemode=UseNone,citecolor=darkgreen,linkcolor=black,urlcolor=BrickRed,pagebackref]{hyperref}
\usepackage[caption=false,font=footnotesize]{subfig}
\usepackage{amsfonts}
\usepackage[utf8]{inputenc}
\usepackage[T1]{fontenc}
\usepackage{textcomp}
\usepackage{arydshln}
\usepackage{multirow}
\usepackage{balance}
\usepackage{tikz}
\usepackage{booktabs}
\usepackage{multirow}
\usepackage{siunitx}
\usetikzlibrary{arrows.meta, positioning}

\renewcommand{\thefigure}{\arabic{figure}}

\definecolor{note}{rgb}{0.1,0.1,1}
\definecolor{rephase}{rgb}{0.15,0.7,0.15}
\definecolor{bag}{rgb}{0.6,0.6,0.2}

\makeatletter
\renewcommand*\env@matrix[1][c]{\hskip -\arraycolsep
  \let\@ifnextchar\new@ifnextchar
  \array{*\c@MaxMatrixCols #1}}
\makeatother

\newcommand{\transpose}{\mathsf{T}}

\makeatletter
\newcommand{\mathleft}{\@fleqntrue\@mathmargin0pt}
\newcommand{\mathcenter}{\@fleqnfalse}
\makeatother

\usepackage{xcolor}
\usepackage{bm}

\title{ Robust Trajectory Tracking of Autonomous Surface Vehicle via Lie Algebraic Online MPC}

\author{Yinan Dong, Ziyu Xu, Tsimafei Lazouski, Sangli Teng, and Maani Ghaffari
\thanks{Y.~Dong, Z.~Xu, T.~Lazouski and M.~Ghaffari are with the University of Michigan, Ann Arbor, MI 48109, USA. }
\thanks{S.~Teng is with the University of California, Berkley, CA 94709, USA.}
}

\begin{document}

\maketitle
\thispagestyle{empty}
\pagestyle{empty}
\newcommand{\reviserbt}[1]{{\color{black}{\emph{#1}}}}

\begin{abstract}
Autonomous surface vehicles (ASVs) are influenced by environmental disturbances such as wind and waves, making accurate trajectory tracking a persistent challenge in dynamic marine conditions. In this paper, we propose an efficient controller for trajectory tracking of marine vehicles under unknown disturbances by combining a convex error-state MPC on the Lie group augmented by an online learning module to compensate for these disturbances in real time. This design enables adaptive and robust tracking control while maintaining computational efficiency. Extensive evaluations in the Virtual RobotX (VRX) simulator and real-world field experiments demonstrate that our method achieves superior tracking accuracy under various disturbance scenarios compared with existing approaches.
\end{abstract} 

\section{Introduction}
The autonomous surface vehicle (ASV) has played an increasingly important role in the application of marine robots, undertaking a wide range of tasks such as hydrographic surveying, environmental monitoring, and water-quality sampling~\cite{zereik2018challenges}.
For surface vehicles, the time-varying hydrodynamic coefficients and external disturbances pose an obstacle to the controller's trajectory-tracking ability, especially for underactuated ASV~\cite{liu2016unmanned}. For example, winds, waves, currents, and inaccurate dynamics modeling can all disturb accurate tracking trajectory~\cite{gao2024dynamic}, which can lead to undesired consequences such as grounding, collision risks, or potential structural damage to the hull, thereby degrading reliability and operational safety.

Classical controllers like PID have been applied on marine robots~\cite{liu2016unmanned}, however, their performance is constrained under strong nonlinearities and time-varying disturbances. In contrast,
 Model Predictive Control (MPC) has demonstrated effectiveness in trajectory tracking, particularly in more complex environments~\cite{cho2020efficient}~\cite{veksler2016dynamic}. However, the non-linear hydrodynamics characteristics mean that the computation load of real-time MPC is heavy, resulting in a long response time to control signals, and a low control rate~\cite{sarda2016station}. Moreover, small computing units lack the capacity to handle such heavy workloads, while large ones consume excessive space and energy, making them impractical for a compact ASV.

Multiple approaches have been put forward to alleviate the computational burden. In~\cite{liu2020computationally}, MPC with projection neural network performs efficient constrained optimization problems with parallel computational capability. In~\cite{annamalai2015robust}, adaptive MPC
leverages multiple approximated linear models to simplify the calculation.
Another approach  uses geometric control on Lie groups to exploit problem symmetries~\cite{ghaffari2022progress} ~\cite{teng2026lieipm}. Lie-algebraic MPC was introduced in ~\cite{teng2022error}, and subsequently applied
to ASVs~\cite{jang2023convex}. Lie-MPC reduces computation time, making online learning for non-linear hydrodynamics applicable to ASV.
\begin{figure}
    \centering
    \includegraphics[height=0.75\columnwidth]{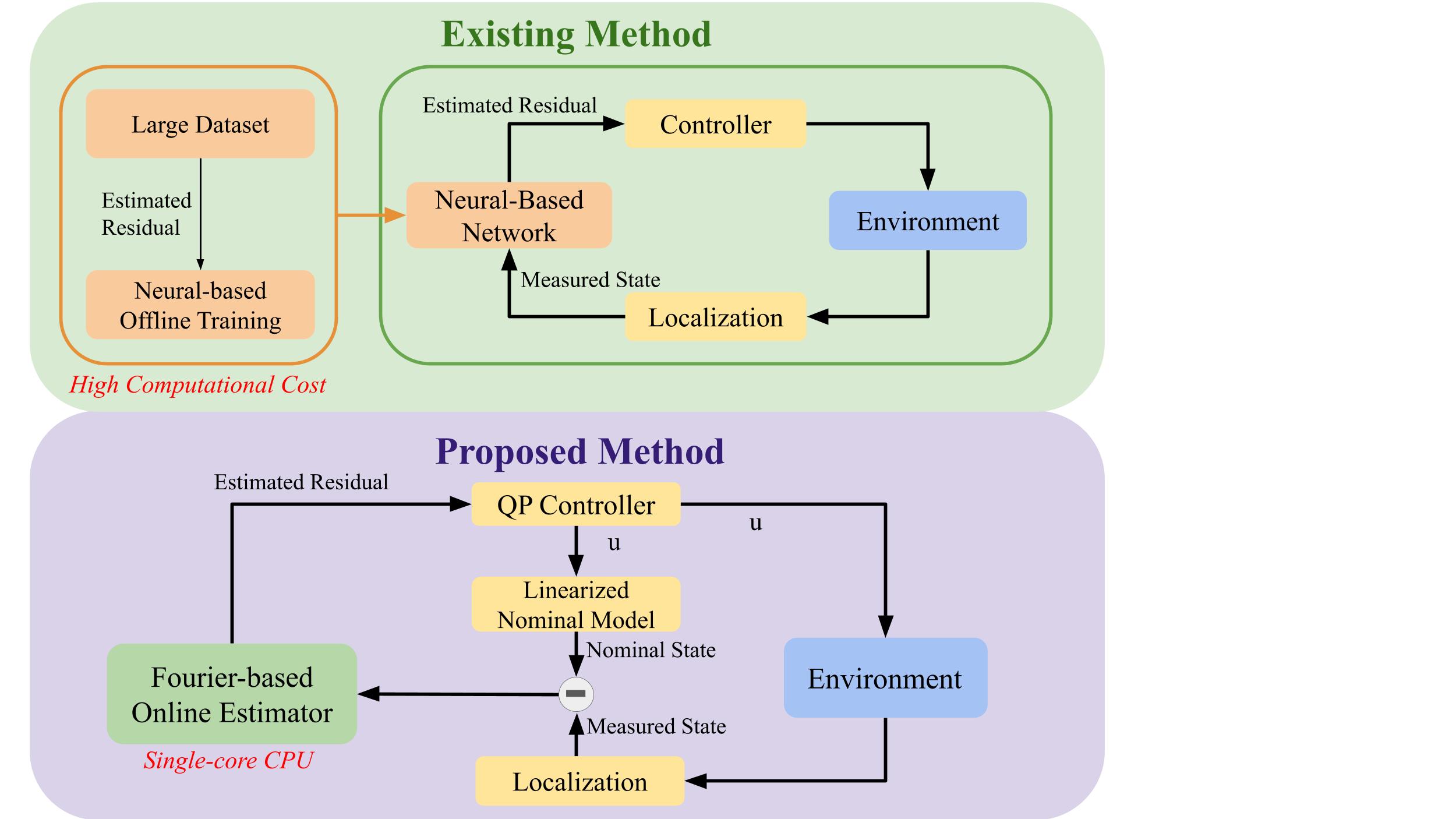}
    \caption{Our method employs Fourier features to estimate the residual in an online manner and can be implemented efficiently on a single-core CPU. In contrast, prior online approaches rely on neural networks that require pre-training on large datasets.}
    \label{fig:firstfig}
\end{figure}

In parallel, several online learning methods have been proposed. 
In~\cite{tu2019adaptive}, a neural network–based adaptive module is incorporated into a model‐based controller to compensate for large wave forces. In~\cite{o2022neural}, the Deep Neural Network (DNN) is first trained offline to learn the model dynamics, then it is updated online to fit unknown disturbances. Similarly, in~\cite{peng2023online}, a DNN is used to learn the system dynamics, and an Extended State Observer (ESO) measures error and estimates the residual dynamics used by the kinematic control law.

However, these neural‐network approaches generally require GPU resources for offline pre‐training on large datasets, which increases computational cost.
Moreover, like in~\cite{peng2023online}, the online-learning pipeline is used to directly change the system dynamics, making it difficult to generalize across different system setups. 
In~\cite{zhou2025simultaneous}, all disturbances are unified into a single locally Lipschitz term within the MPC dynamics and estimated through feature-based modeling. While this approach has been demonstrated on aerial vehicles and shown to reduce computational burden, its applicability to ASVs and its interaction with a Lie-MPC controller remain unexplored.
In particular, although the formulation in~\cite{zhou2025simultaneous} is general for control-affine systems, it relies on solving a nonlinear MPC (NMPC) problem online, which can impose significant computational demands. This becomes especially critical for ASV platforms, where onboard computation must be shared among localization and perception modules.

This work builds upon the convex Lie-MPC framework introduced in~\cite{jang2023convex}, which formulates trajectory tracking as a convex error-state MPC problem on the Lie algebra. Compared with~\cite{jang2023convex}, the linearized error-state dynamics are augmented with a residual term $\delta_t$ to capture higher-order linearization errors and unmodeled hydrodynamic disturbances. We also integrate a Fourier-based online residual estimator to enable real-time disturbance compensation, and an offline bi-level Fourier feature extraction framework to jointly optimize Fourier frequencies and coefficients for informative disturbance representation.

To mitigate the research gap, the main contributions of this work are as follows:

\begin{enumerate}
    \item \textbf{A real-time online learning framework for MPC on ASVs.} 
    We propose a novel integration of Fourier-based online learning with convex Lie algebraic MPC that enables real-time disturbance compensation using only a single CPU core. 

     \item \textbf{Bi-level feature extraction for efficient disturbance representation.}  
    We develop a bi-level Fourier-based feature extraction method that identifies and retains the dominant components of disturbance dynamics.

    \item \textbf{Validation across simulation and real-world domains.}  
    The proposed controller is evaluated in a physics-based simulator and in real-world experiments. 
\end{enumerate}

The remainder of this paper is organized as follows. Section II provides the mathematical preliminaries. Section III presents the details of our proposed method. Experiments are presented in Section IV. Finally, Section V concludes the paper and discusses future studies.

\section{Preliminaries and Mathematical Formulation}
This section provides a brief overview of the necessary mathematical background used in the developed approach.
\begin{figure*} [t]
    \centering
    \includegraphics[width=1.5\columnwidth]
    {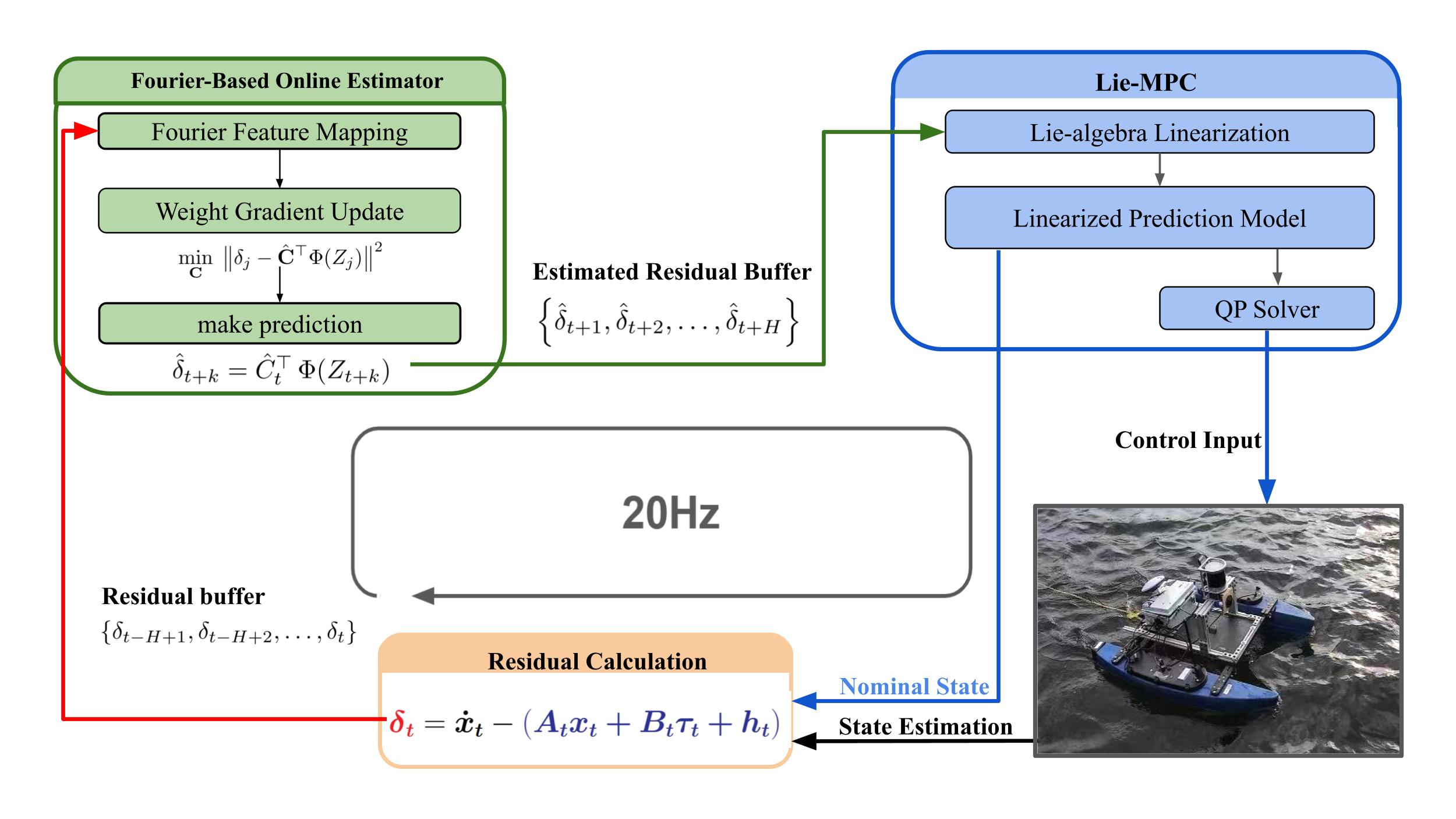}
    \caption{This figure illustrates the overall pipeline. At each control cycle, a residual term is first computed by comparing the measured state derivative with the nominal dynamics model. A sliding buffer containing the most recent residual samples serves as the input to the Fourier-based online estimator. Then, the estimator updates the weight matrix to approximate this residual-history buffer and predicts a full horizon of future residual terms. These predicted residuals are then injected into the Lie-MPC, allowing the controller to account for upcoming disturbances over the entire planning horizon. In this manner, the proposed framework achieves real-time disturbance compensation without any weight pre-training and operates efficiently on resource-limited ASV platforms.}
    \label{fig:pipeline}
    \vspace{-4mm}
\end{figure*}
\subsection{Dynamics on Lie Group}
This section briefly introduces the rigid body dynamics on the Lie group $\mathrm{SE}(3)$. For more backgrounds on matrix Lie groups, refer to \cite{hall2013lie, sola2018micro, bullo2019geometric}.

Let $\mathcal{G}$ be an $n$-dimensional matrix Lie group with its associated Lie algebra $\mathfrak{g}$. We define the following isomorphism
\begin{equation}
    (\cdot)^\wedge:\mathbb{R}^n \rightarrow \mathfrak{g},
\end{equation}
that maps an element in the vector space $\mathbb{R}^n$ to the tangent space of the matrix Lie group at the identity. Its inverse mapping is denoted as $(\cdot)^\vee:\mathfrak{g}\rightarrow\mathbb{R}^n$. Then, for any $\phi \in \mathbb{R}^{n}$, we define the exponential map as
\begin{equation}
    \exp(\cdot):\mathbb{R}^{n} \rightarrow \mathcal{G},\ \ \exp(\phi)=\operatorname{exp_m}({\phi}^\wedge),
\end{equation}
where $\operatorname{exp_m}(\cdot)$ denotes the matrix exponential.

To describe the 3D rigid body motion of the vehicle, we represent the rotation by matrix $R \in \mathrm{SO}(3)$ and the position vector by $p \in \mathbb{R}^3$ and have the full state as:
\begin{equation}
    X =
    \begin{bmatrix}
        R & p \\
        0 & 1
    \end{bmatrix} \in \mathrm{SE}(3).
\end{equation}
The tangent space at the identity, $\mathfrak{se}(3)$, consists of twist $\xi \in \mathbb{R}^6$, which is defined as the concatenation of the angular velocity $\omega$ and the linear velocity $v$ in the body frame:
\begin{equation}
    \xi = 
    \begin{bmatrix}
        \omega \\
        v
    \end{bmatrix}, \quad
    \xi^\wedge =
    \begin{bmatrix}
        \omega^\wedge & v \\
        0 & 0
    \end{bmatrix}
    \in \mathfrak{se}(3).
\end{equation}
The adjoint map of $\mathrm{SE}(3)$ has the matrix representation:
\begin{equation}
    \mathrm{Ad}_{X} =
    \begin{bmatrix}
        R & 0 \\
        p^\wedge R & R
    \end{bmatrix},
    \qquad
    \mathrm{ad}_{\xi} =
    \begin{bmatrix}
        \omega^\wedge & 0 \\
        v^\wedge & \omega^\wedge
    \end{bmatrix},
    \label{eq:adjoint}
\end{equation}
where $\mathrm{Ad}_{X}$ describes the change of frame of Lie algebra, and $\mathrm{ad}_{\xi}$ is defined by $\mathrm{ad}_{\xi}\eta = [\xi, \eta]$ with $[\cdot, \cdot]$ the Lie bracket.

\subsection{Dynamics of ASV in Fossen's model}
We will use a 6-DOF motion equations based on Fossen's model~\cite{fossen1995nonlinear} to describe the ASV, with the prescribed Lie group's ordering of twist vector $\xi=[\omega,v]$. The dynamics are represented in the body frame following the convention of Euler Poincare equation for single rigid body:
\begin{align}
M_{RB}\dot{\xi}
&+ C_{RB}(\xi)\xi
+ M_{AM}\dot{\xi}_r
+ C_{AM}(\xi_r)\xi_r \notag\\
&+ D(\xi_r)\xi_r
+ g(\eta)
= \tau_c + \tau_{disturb}^{body}
\label{eq:Fossen_dynamics}
\end{align}


\noindent with $M_{(\cdot)}$ the mass matrix, $C_{(\cdot)}$ the Coriolis matrix. The subscript $RB$ refers to the rigid-body
contribution, while $AM$ means ``added mass'', which represents the inertia of fluid accelerated by the vehicle motions. $D(\xi_r)$ is the hydrodynamic damping (drag) matrix. $g(\eta)$ is the hydrostatic restoring force and moment vector, and here we assume 
the vehicle to be neutrally buoyant, so the hydrostatic term $g(\eta)$ becomes negligible.
 $\xi_r = \xi - \xi_c$ is the relative twist with respect to the ocean current $\xi_c$. $\tau_c$ is the six-dimensional control input from the propulsion system, $\tau_{disturb}^{body}$ is the six-dimensional external disturbance transformed in the body frame.


The mass matrix corresponding to $\xi$ is given by
\begin{equation}
M =
\begin{bmatrix}
M_{11} & M_{12} \\
M_{21} & M_{22}
\end{bmatrix},
\quad
M_{ij} \in \mathbb{R}^{3\times3},
\label{eq:MRB}
\end{equation}
where the submatrices are defined as
$M_{11}$ is a positive definite inertia matrix in body frame, 
$M_{12} = M_{21}^\top = -m\, r_g^{\wedge}$,
and $M_{22} = m I_3$,
where $r_g = [x_g, y_g, z_g]^\transpose$ denotes the position vector of the
center of gravity expressed in the body-fixed frame.  $(\cdot)^{\wedge} :
\mathbb{R}^3 \rightarrow \mathfrak{so}(3)$ denotes the wedge operator that maps
a vector to its corresponding skew-symmetric matrix.
The Coriolis matrix is given by
\begin{equation}
\begin{aligned}
C(\xi) =
\begin{bmatrix}
-\left(M_{11}\omega + M_{12}v\right)^{\wedge} & -\left(M_{21}\omega + M_{22}v\right)^{\wedge} \\
-\left(M_{21}\omega + M_{22}v\right)^{\wedge} & 0
\end{bmatrix}
\end{aligned}
\label{eq:Coriolis}
\end{equation}
which is a skew-symmetric matrix representing the ``inertial forces'' in the vehicle dynamics.
The damping matrix is given by
\begin{equation}
\begin{aligned}
D(\xi_r) &= 
- \, \mathrm{diag}\!\left(
\begin{bmatrix}
K_p, \; M_q, \; N_r, \; X_u, \; Y_v, \; Z_w
\end{bmatrix}^\transpose
\right) \\
&\quad
- \, \mathrm{diag}\!\left(
\begin{bmatrix}
K_{|p|p}|p|, \; M_{|q|q}|q|, \; N_{|r|r}|r|, \\[6pt]
X_{|u|u}|u|, \; Y_{|v|v}|v|, \; Z_{|w|w}|w|
\end{bmatrix}^\transpose
\right)
\end{aligned}
\label{eq:Damping}
\end{equation}                                 
The damping terms model hydrodynamic effects such as potential damping, skin friction, wave drift damping, vortex shedding, and lifting forces.  
Depending on the surge velocity, either linear or nonlinear damping terms dominate.  
For non-coupled motion, a diagonal damping structure can be assumed, simplifying the dynamics.

Combining \eqref{eq:adjoint} with \eqref{eq:Fossen_dynamics}, have the full dynamics
\begin{equation}
M \dot{\xi} = \mathrm{ad}^\transpose_{\xi} M \xi + F 
= -C(\xi)\xi - D(\xi)\xi + \tau,
\end{equation}
where the ocean current velocity $\xi_c$ is assumed to be unknown, and the corresponding current-induced effects are absorbed into the disturbance term for simplicity.

\begin{equation}
\begin{bmatrix}
\dot{R} & \dot{p} \\
0 & 0
\end{bmatrix}
=
\begin{bmatrix}
R & p \\
0 & 1
\end{bmatrix}
\begin{bmatrix}
\omega^{\wedge} & v \\
0 & 0
\end{bmatrix}.
\end{equation}

\noindent
with $F \in \mathfrak{se}^*(3) $ the external force in the body frame.

\section{Methods}
In this section, we present the proposed framework that integrates an error-state convex Lie-MPC with an online Fourier-based learning module for disturbance compensation. 

\subsection{Error-State Convex MPC}
We adopt a convex error-state MPC formulation based on the Lie algebra representation, following the framework introduced in~\cite{jang2023convex}.

Let the desired trajectory be $X_{d,t} \in \mathrm{SE}(3)$, and define the left-invariant error \cite{barrau2016invariant} on matrix Lie group as:
\begin{equation}
    \Psi = X_{d,t}^{-1} X_t, \quad 
    \dot{\Psi} = \Psi (\xi_t - Ad_{\Psi_t^{-1}}\xi_{d,t})^\wedge.
\end{equation}

Using the first-order approximation $\Psi_t \approx I + \psi_t^\wedge$, where $\psi_t \in \mathbb{R}^6$ is the error vector in the Lie algebra obtained via the logarithmic map of $\Psi_t$, we obtain the linearized error dynamics in Lie algebra:
\begin{equation}
    \dot{\psi}_t = -ad_{\xi_{d,t}}\psi_t + \xi_t - \xi_{d,t}.
\end{equation}

Linearizing the hydrodynamic model around $\bar{\xi} = \xi_{d,t}$ yields:
\begin{equation}
    M \dot{\xi} = H_t \xi + b_t +M\delta_t^{d} + \tau,
\end{equation}
where $H_t$ and $b_t$ are the Jacobian and residual terms derived from the first-order expansion of $C(\xi)$ and $D(\xi)$, while $\delta_t^{d}$ is derived from external hydrodynamic disturbance.

Define the system state $x_t = [\psi_t^\transpose, \xi_t^\transpose]^\transpose$, and the linearized hydrodynamic model becomes:
\begin{equation}
    \label{eq:system_dynamics}
    \dot{x}_t = A_t x_t + B_t \tau + h_t + \delta_t,
\end{equation}
with each component defined by:
\begin{equation}
\begin{aligned}
    A_t &= 
    \begin{bmatrix}
        -ad_{\xi_{d,t}} & I \\[4pt]
        0 & M^{-1}H_t
    \end{bmatrix}, \quad
    B_t =
    \begin{bmatrix}
        0 \\[4pt]
        M^{-1}
    \end{bmatrix}, \\[10pt]
    h_t &=
    \begin{bmatrix}
        -\xi_{d,t}\\[4pt]
       M^{-1} b_t 
    \end{bmatrix}, \quad
    \delta_t =
    \begin{bmatrix}
        \delta_t^{e} \\[4pt]
        \delta_t^{d}
    \end{bmatrix}.
\end{aligned}
\label{eq:system_matrices}
\end{equation}

Here,
    $\delta_t^{e}$ represents the higher-order linearization residuals 
    in the error-state dynamics, capturing higher-order terms 
    neglected in the first-order approximation around $\xi_{d,t}$,
    $\delta_t^{d}$ denotes the external disturbances and unmodeled hydrodynamic effects.

For cost function, we first define the tracking output:
\begin{equation}
    y_t = 
    \begin{bmatrix}
        \psi_t \\ \dot{\psi}_t
    \end{bmatrix}
    = G_t x_t - d_t,
\end{equation}
where
\begin{equation}
    G_t =
    \begin{bmatrix}
        I & 0 \\ -ad_{\xi_{d,t}} & I
    \end{bmatrix}, \quad
    d_t =
    \begin{bmatrix}
        0 \\ \xi_{d,t}
    \end{bmatrix}.
\end{equation}

Then the cost function is formulated as:
\begin{equation}
    J = y_N^\transpose P y_N + \sum_{k=1}^{N-1} (y_k^\transpose Q y_k + \tau_k^\transpose R \tau_k),
\end{equation}
with $P$, $Q$, and $R$ the positive definite weighting matrices.

After discretization with step $\Delta t$, the MPC optimization problem becomes:
\begin{equation}
\begin{aligned}
    \min_{\tau_k} \quad & y_N^\transpose P y_N + \sum_{k=1}^{N-1}(y_k^\transpose Q y_k + \tau_k^\transpose R \tau_k) \\
    \text{s.t.} \quad & x_{k+1} = A_k x_k + B_k \tau_k + h_k+ \delta_k, \\
    & x_0 = x_{\text{init}}, \quad \tau_k \in \mathcal{T}_k,
\label{eq:mpc_problem}
\end{aligned}
\end{equation}
where $A_k = I + A_{t_k}\Delta t$, $B_k = B_{t_k}\Delta t$, $h_k = h_{t_k}\Delta t$, and $\delta_k =\delta_{t_k}\Delta t$.  
The resulting QP can be efficiently solved using the OSQP solver.

\begin{algorithm} 
\caption{Fourier Online MPC}
\label{alg:mpc_rff_update}
\begin{algorithmic}[1]
\Require Total timesteps $T$; MPC prediction horizon $N$;
feature map $\Phi(\cdot)$;
step size $\eta$;
regularizer $\lambda$
\State $x_1 \gets [\psi_1, \xi_1]^\top$ 
\Comment{Initialize system state}

\State  $\hat{C}_1 \gets 0$ 
\Comment{Initialize weight matrix}

\State $\hat{\delta}_{0|0} \gets 0$ 
\Comment{Initialize predicted residual}
\For{$t = 1,\ldots,T$}                                    
    \State
    {\small
    $\{\hat{\delta}_{t+k|t}\}_{k=1}^{N}
    \;\gets\;
    \{\hat{C}_t^\top \Phi(Z_{t+k|t})\}_{k=1}^{N}$}\Comment{Predict Residual}
    \State
    $\tau_{t}
    \;\gets\;
    \mathrm{MPC}\!\big(
    x_t,\,
    \{\hat{\delta}_{t+k}\}_{k=1}^{N}
    \big)$
    \Comment Solve~\eqref{eq:mpc_problem}
    \State 
    $\delta_{t} = \dot{x}_{t} - (A_{t} x_{t} + B_{t} \tau_{t} + h_{t})$
    \Comment Estimate residual
    \State
    $\ell_t\gets~\eqref{eq:buffered_loss}$\Comment Compute loss
    \State
    $\hat{C}_{t+1} \gets \hat{C}_t - \eta \nabla_{\hat{C}_t} \ell_t$
    \Comment Update weights
\EndFor
\end{algorithmic}
\end{algorithm}

\subsection{Fourier-based Online Learning}
At each time step $t$, we estimate the residual term $\delta_t$ by comparing the measured and nominal state derivatives:
\begin{equation}
    \delta_t = \dot{x}_{t} - (A_t x_t + B_t \tau_t + h_t),
\label{eq:residual_calculation}
\end{equation}
where $\dot{x}_{t} = (x_{t+1} - x_t)/\Delta t$ is obtained from discrete-time measurements  using GPS and IMU fusion\cite{singh2025gps}, and $(A_t x_t + B_t \tau_t + h_t)$ is the nominal model from~\eqref{eq:system_dynamics}. 
The residual $\delta_t = [\delta_t^{e}, \delta_t^{d}]^\top$ captures both internal linearization errors and external disturbances.

We approximate the residual by a Fourier basis evaluated on an input that concatenates the state $x_t$ and time $t$:
\begin{equation}
        Z_t = (x_t, t), \qquad 
    \hat{\delta}_t = {C}^\top \,\Phi(Z_t),
\end{equation}
where $\Phi(\cdot):\mathbb{R}^{12} \times \mathbb{R} \rightarrow \mathbb{R}^{d}$ is the Fourier feature map that maps the time-dependent robot state to a $d$-dimensional feature. The coefficient matrix ${C}$ is updated online using samples stored in a sliding buffer $\mathcal{B}$.

We use the most recent samples to compute a buffered loss with a temporal smoothness regularizer:
\begin{equation}
\label{eq:buffered_loss}
    \ell_t \;=\; \frac{1}{W}\!\sum_{j=t-W+1}^{t}
    \Big(\,\|\delta_j - \hat{\delta}_j\|^2
    + \lambda\,\|\hat{\delta}_j - \hat{\delta}_{j-1}\|^2\Big),
\end{equation}
where W denotes the length of the sliding window. We set $\hat{\delta}_{0}\!= 0$ for the smoothness term when $j=1$.
Rather than solving for the exact minimizer, we perform gradient
step at each control cycle:
\begin{equation}
\hat{C}_{t+1} \gets \hat{C}_t - \eta \nabla_{\hat{C}_t} \ell_t
\end{equation}

The use of Fourier features provides a flexible and expressive representation for approximating time-varying and state-dependent disturbances. In particular, it enables the model to capture periodic variations in the residual dynamics while maintaining a linear parameterization with respect to the coefficients, thereby preserving the convexity of the optimization problem in Eq.~(\ref{eq:mpc_problem}).

The proposed online learning formulation is computationally efficient, the coefficient update performs fixed gradient step on
the buffered least-squares objective. This makes it suitable for real-time implementation alongside the MPC controller.

\subsection{Offline Bi-level Feature Extraction}
Inspired by the Fourier-based nature of wave and wind disturbances~\cite{lale2024falcon}, the features we use are all Fourier-based. However,
directly using a large number of random Fourier features for disturbance estimation causes many of them fail to capture the dominant frequency components and spatial correlations of the residual term. To address this issue, we perform feature extraction to identify and retain components that contribute significantly to the residual representation. This reduces both the computational cost and the risk of overfitting while preserving model accuracy. 

Each feature is represented as a sine–cosine pair obtained from a single variable $z_{t,i}$ with a trainable frequency $f_j$:
\begin{equation}
    \Phi(Z_t) =
    \big[ \sin(f_j\, z_{t,i}), \; \cos(f_j\, z_{t,i}) \big]_{i=1,\dots,d;\, j=1,\dots,m}
\end{equation}
where $d$ is the state dimension and $m$ is the number of selected frequencies.

\begin{algorithm}
\caption{Offline Bi-level Feature Extraction}
\label{alg:feature_extraction}
\begin{algorithmic}[1]

\Require Dataset $\{(Z_t, \delta_t)\}_{t=1}^{T}$;
initial amplitudes $\hat{C}$;
initial frequency set $\{f_j\}_{j=1}^M$;
feature map $\Phi(\cdot)$;
inner steps $n_{\mathrm{inner}}$;
epochs $N$;
learning rate $\eta$;
regularization $\lambda$

\Statex \textbf{Output:} Optimized Fourier feature map $\Phi(\cdot)$

\For{$\text{epoch} = 1$ \textbf{to} $N$}

    \State Freeze $\{f_j\}$; unfreeze $\hat{C}$
    \Comment Inner loop
    \For{$k = 1$ \textbf{to} $n_{\mathrm{inner}}$}
        \ForAll{$(Z_t, \delta_t)$ \textbf{in dataset}}
            \State $\hat{\delta}_t \gets \hat{C}^\top \Phi(Z_t)$
            {\small\State $L_{\mathrm{inner}} \gets \|\hat{\delta}_t - \delta_t\|_2^2 + \lambda \|\hat{C}\|_1$}
            \Comment Compute loss
            \State $\hat{C} \gets \hat{C} - \eta \nabla_{\hat{C}} L_{\mathrm{inner}}$
            \Comment Update weights
        \EndFor
    \EndFor

    \State Freeze $\hat{C}$; unfreeze $\{f_j\}$
    \Comment Outer loop
    \ForAll{$(Z_t, \delta_t)$ \textbf{in dataset}}
        \State $\hat{\delta}_t \gets \hat{C}^\top \Phi(Z_t)$
        {\small\State $L_{\mathrm{outer}} \gets \|\hat{\delta}_t - \delta_t\|_1$}
        \Comment Compute loss
        \State $\{f_j\} \gets \{f_j\} - \eta \nabla_{f_j} L_{\mathrm{outer}}$
        \Comment Update weights
    \EndFor

\EndFor

\State $\Phi(Z_t)
\;\xleftarrow[\text{30 features}]{\text{PCA}}
\;\Phi(Z_t)$
\Comment Extract Features

\end{algorithmic}
\end{algorithm}

Let $\{(Z_t,\delta_t)\}_{t=1}^{T}$ denote the dataset, where $T$ is the total number of samples.We use a bi-level optimization to extract features, which alternates between amplitude and frequency updates. The problem is formulated as
\begin{equation}
\min_{\{f_j\}} \;
   \frac{1}{T}\sum_{t=1}^{T} \ell\!\Big(\delta_t,\;
   \hat{C}(f)\,\Phi(Z_t)\Big).
\end{equation}
where
\begin{equation}
   \hat{C}(f) \;=\; \arg\min_{C}
   \frac{1}{T}\sum_{t=1}^{T} \ell\!\big(\delta_t,\,C\,\Phi(Z_t)\big)
   + \lambda \|C\|_1 .
\end{equation}
with $\ell(\cdot,\cdot)$ denoting the L2 prediction loss.  
The inner optimization loop optimizes the amplitude parameters $C$ with an $\ell_1$ regularization that promotes sparsity, while the outer optimization loop adjusts the frequencies $\{f_j\}$ to minimize the overall prediction error.

The inner optimization is a convex problem with respect to $C$, which guarantees a unique global optimum for fixed frequencies. We solve the overall bilevel objective using alternating minimization: fix $\{f_j\}$ and update $C$ (inner loop), then fix $C$ and update $\{f_j\}$ (outer loop). 
This alternating scheme adaptively learns both the most informative frequencies and their corresponding amplitudes. 
Note that while the inner problem is convex, the outer optimization over the frequency parameters $\{f_j\}$ is generally non-convex, due to the nonlinear dependence of the Fourier feature map on the frequencies. But as mentioned in the Proposition 1 of \cite{zhou2025simultaneous}, the performance is within a regret bound even with suboptimal feature selection. 
Even in the presence of Gaussian noise in the disturbance data, the convexity of the inner layer ensures convergence to a stable global minimum. 
Non-Gaussian noise may lead to suboptimal frequency selection in the outer optimization. However, $\ell_2$ fitting remains a consistent estimator under broad noise distributions and is computationally favorable for the convex inner update.
The implementation is detailed in Algorithm \ref{alg:feature_extraction}.

\section{Experiments}

\begin{figure}[b]
    \centering
    \includegraphics[width=0.9\columnwidth]
    {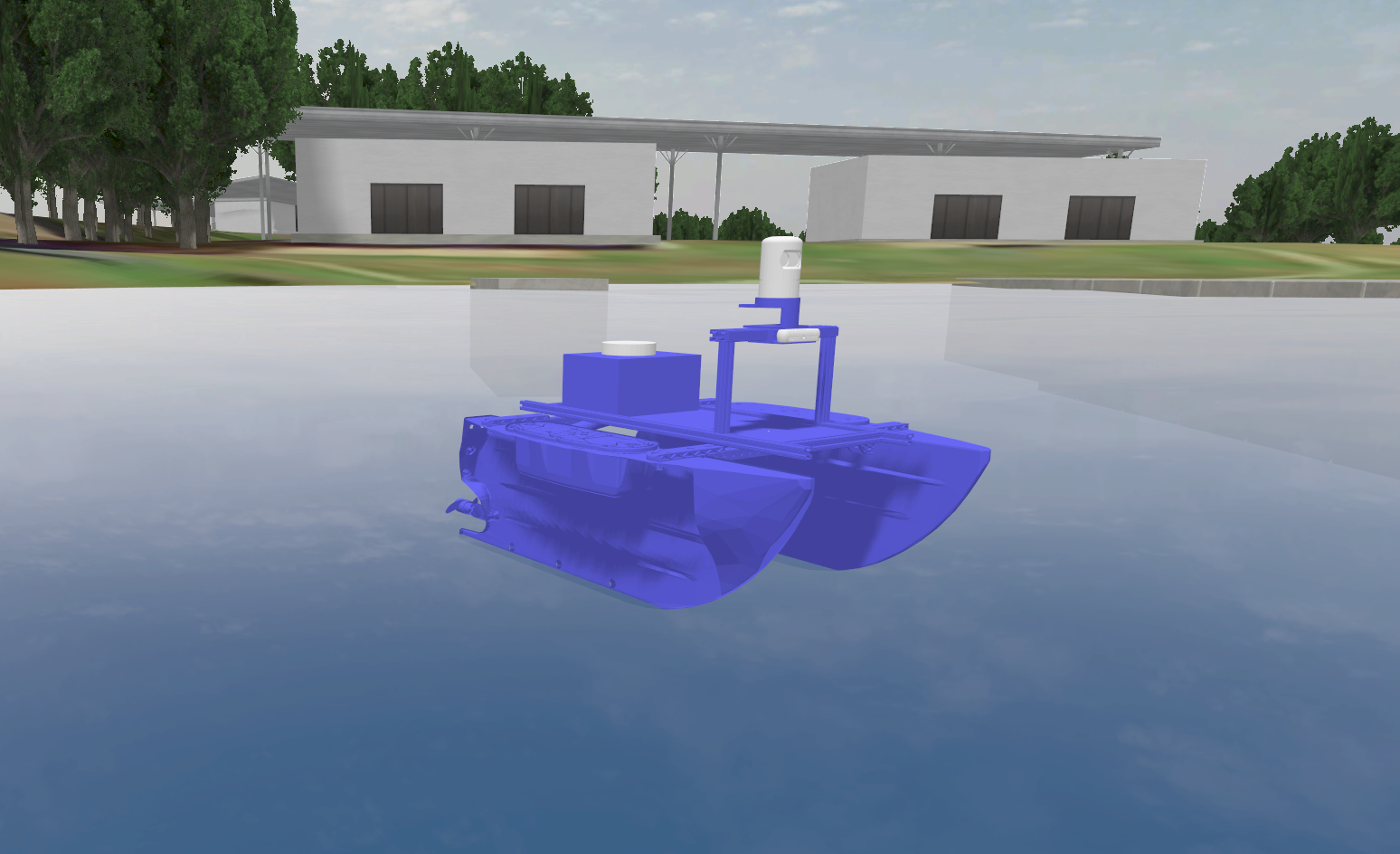}
    \caption{The VRX environment.}
    \label{fig:sim_histogram}
\end{figure}

\begin{figure*} 
    \centering
    \includegraphics[width=2\columnwidth]
    {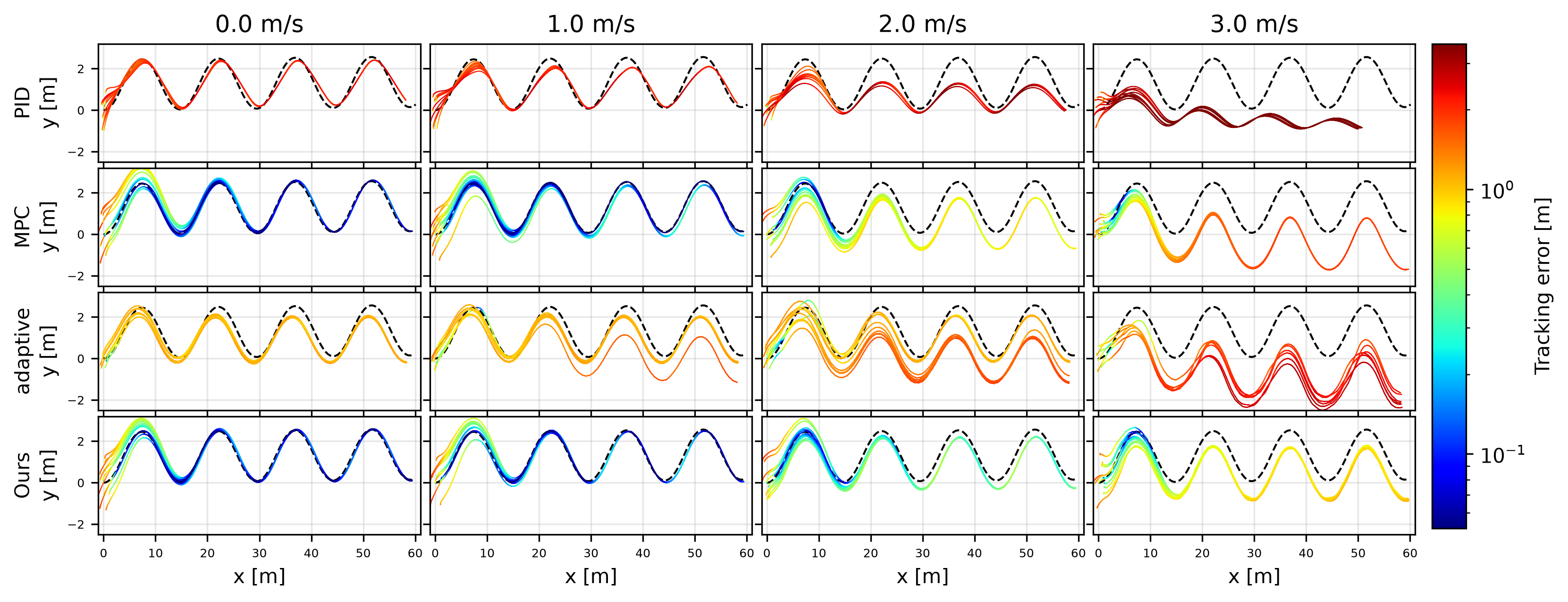}
    \caption{Tracking performance of baselines and our method under different wind conditions in the VRX simulator.}
    \label{fig:vrx_trajectory_zigzag}
\end{figure*}

\begin{figure*} 
    \centering
    \includegraphics[width=2\columnwidth]
    {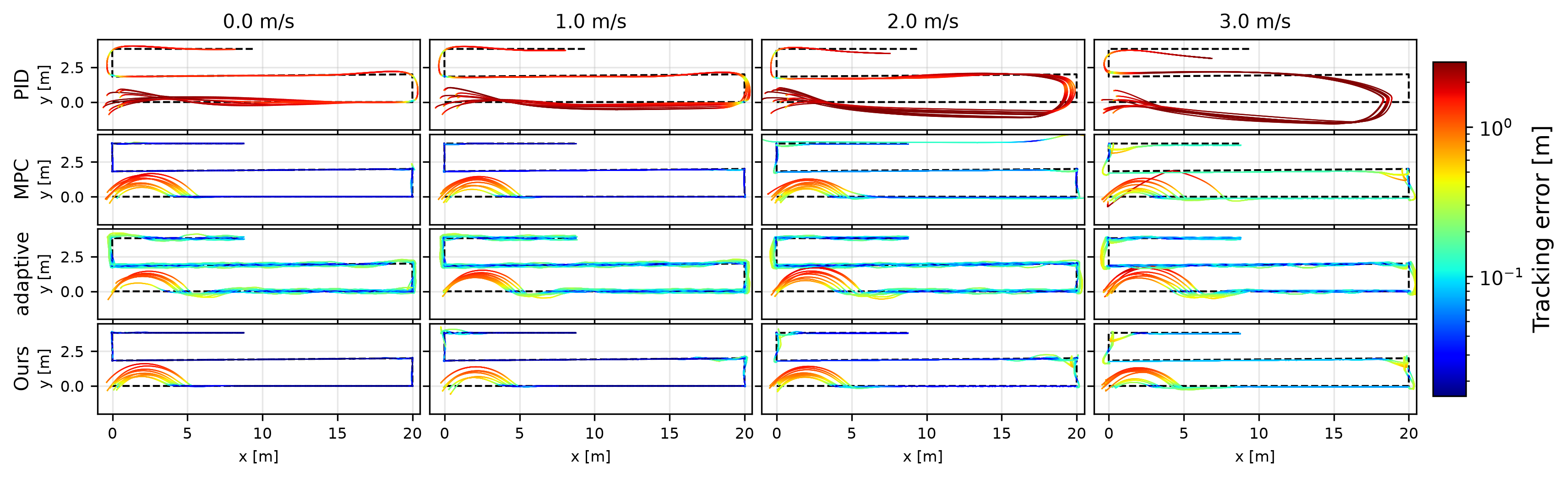}
    \caption{Tracking performance of baselines and our method under different wind conditions in the VRX simulator.}
    \label{fig:vrx_trajectory_lawnmower}
\end{figure*}

\subsection{Experiment Setup}
\subsubsection{Repetition Protocol}
This section involves tests under two scenarios: (i) the Virtual RobotX (VRX) simulator, and (ii) a real-world river experiment under wind. Tests in scenarios (i) are each repeated 10 times, with an initial position offset
$\Delta \mathbf{p}_0 = [\,1.0\sqrt{u}\cos\theta,\; 1.0\sqrt{u}\sin\theta,\; 0\,]^\top \mathrm{m}$,
where $\theta \sim \mathcal{U}(0,2\pi)$ and $u \sim \mathcal{U}(0,1)$. 

\subsubsection{Controller Tuning}
To reduce the coupling between observer and controller, we first approximate the dynamics with an
LQR approximation. Given weighting matrices $Q$, $R$, we
compute $P$ from the discrete-time algebraic Riccati equation (DARE)
\begin{equation}
P = A^{\top} P A - A^{\top} P B (R + B^{\top} P B)^{-1} B^{\top} P A + Q,
\label{eq:DARE}
\end{equation}
and set the stabilizing gain
\begin{equation}
K = (R + B^{\top} P B)^{-1} B^{\top} P A.
\label{eq:K}
\end{equation}
We tune weighting matrices $Q$ and $R$ based on eigenvalues of $A - B K$, avoiding complex conjugate pairs and ensuring the norm remains well below unity.

\subsubsection{Baselines}
We compare the performance of our method with those baseline methods: (1) \textbf{Nominal Lie-MPC}: the Lie-MPC framework without online learning~\cite{jang2023convex}. (2) \textbf{L1 Adaptive-MPC}: the same MPC framework with an online L1-style disturbance–estimation module to calculate compensation based on residual dynamics as well.~\cite{1657243} (3) \textbf{PID}: a two-layer trajectory-tracking controller consisting of an outer kinematic feedback loop:
{
\begin{equation} \begin{aligned} v_x^{\mathrm{ref}} &= v_x^{\mathrm{ff}} + k_{xy}e_x^b,\\ \omega_z^{\mathrm{ref}} &= \omega_z^{\mathrm{ff}} + k_{\psi}e_{\psi} + k_{\mathrm{lat}}e_y^b, \end{aligned} \end{equation} and an inner loop that computes the surge force and yaw moment as \begin{equation} \begin{aligned} F_x &= \mathrm{PID}_v\!\left( v_x^{\mathrm{ref}}-v_x+\alpha e_x^b \right),\\ M_z &= \mathrm{PID}_{\omega}\!\left( \omega_z^{\mathrm{ref}}-\omega_z+\beta e_y^b \right), \end{aligned} \end{equation} where $e_x^b$ and $e_y^b$ are the longitudinal and lateral position errors in the body frame, respectively, and $e_{\psi}$ is the yaw error. The differential thrust command is obtained as \begin{equation} u = \begin{bmatrix} u_r\\ u_l \end{bmatrix} = \begin{bmatrix} \frac{1}{2}F_x+\frac{M_z}{2L}\\ \frac{1}{2}F_x-\frac{M_z}{2L} \end{bmatrix}. \end{equation}
}
where $u$ denotes the differential thrust command. 

\subsubsection{Features for disturbance estimation}
After collecting disturbance and corresponding state variable from 10 simulation runs in VRX simulator. We employ Algorithm~\ref{alg:feature_extraction} to extract 30 features, which are then used in multiple scenarios under various disturbances. The extracted features exhibit strong generalization capability.

\subsubsection{Performance metric}
We use the Root Mean Square Error (RMSE) as the performance metric for planar tracking. Since a random initial position error is introduced in each test, the RMSE is evaluated only after the ASV reaches a steady state to exclude the relatively large error at the beginning of each run. For instance, the RMSE is calculated over the second half of the path. 
For the selected segment $\mathcal{K}$, the trajectory-tracking RMSE for each run is
\[
\mathrm{RMSE} = \sqrt{\frac{1}{|\mathcal{K}|}\sum_{k\in\mathcal{K}} \big\| [x_k,\, y_k]^\top - [x_k^{\mathrm{ref}},\, y_k^{\mathrm{ref}}]^\top \big\|_2^2 } .
\]
The values in the table are the mean and standard deviation of $\mathrm{RMSE}$ across repeated trials.

\subsubsection{Implementation details}

The experiments are conducted using two reference trajectories: 
(i) a zigzag path defined by a time-varying yaw rate 
$\omega_z(t)=0.1\cos(t/100)$ with constant surge speed $v_x=0.5\,\mathrm{m/s}$, 
and (ii) a lawn-mower coverage pattern characterized by repeated straight segments 
($v_x=0.5\,\mathrm{m/s}$) and 90° turns executed at a turning rate 
$\omega_z=0.3\,\mathrm{rad/s}$. 
Both trajectories have a total duration of $128\,\mathrm{s}$.

To incorporate the learned disturbances into the global MPC model, the estimated body-frame disturbance wrench 
$\tau_t^{\text{body}}$ is transformed into the world frame via
\begin{equation}
    \tau_t^{\text{world}} = 
    \begin{bmatrix}
        R(X_t) & 0 \\
        0 & R(X_t)
    \end{bmatrix}
    \tau_t^{\text{body}},
\end{equation}
where $R(X_t)$ is the rotation matrix corresponding to the ASV orientation.
This operation assumes spatially uniform disturbances acting at the center of mass.

The algorithm configuration used across all experiments is summarized in 
Table~\ref{tab:MPC-Configuration}. 

\begin{table}[h]
\centering
\caption{Algorithm Configuration}
\label{tab:MPC-Configuration}
\resizebox{0.95\columnwidth}{!}{
\begin{tabular}{@{}lcccc@{}}
\midrule
 Control Rate & Control Horizon $N$ & step size $\eta$ & Regularizer $\lambda$ \\
\midrule
50Hz & 30 & $10^{-4}$ & $10^{-4}$ \\
\midrule
\end{tabular}
}
\end{table}

\subsection{Simulations in physics-based Virtual RobotX simulator}
\subsubsection{System Setup}
We evaluate algorithm performance using the Virtual RobotX (VRX) simulator, which provides a physics-based marine environment with realistic hydrodynamic and wind effects, as shown in Fig.~\ref{fig:sim_histogram}. In simulations, the wave model and random wind seed are disabled for consistency, while the wind gain is fixed at 1 under the constant-speed option. A URDF model of the real-world ASV is imported into the simulator; the vessel has a mass of 28~kg and is actuated by two fixed parallel propellers with thrust limits of $[-26,50]$~N.

\subsubsection{Results}
As shown in Table~\ref{tab:vrx_tracking_zigzag}, Table~\ref{tab:vrx_tracking_lawnmower}, Fig.~\ref{fig:vrx_trajectory_zigzag}, and Fig.~\ref{fig:vrx_trajectory_lawnmower}, the proposed Online-MPC achieves lower tracking error across all wind conditions and trajectories. For both the zigzag and lawnmower trajectories, Online-MPC outperforms the nominal MPC, with the performance gap widening as wind speed increases, demonstrating improved robustness and disturbance-rejection capability under stronger environmental disturbances.

\begin{table}[h]
\centering
\caption{Trajectory-tracking RMSE for zigzag trajectory in VRX simulator. 
The values in the first row refer to the set wind speed in the simulator.}
\label{tab:vrx_tracking_zigzag}
\begingroup
\setlength{\tabcolsep}{2.5pt}
\renewcommand{\arraystretch}{0.9}
\resizebox{1.00\columnwidth}{!}{
\begin{tabular}{@{}l*{4}{c}@{}}
\toprule
Controller & 0.0 m/s & 1.0 m/s & 2.0 m/s & 3.0 m/s \\
\midrule

PID &
$1.928 \pm 0.013$ &
$1.992 \pm 0.031$ &
$2.588 \pm 0.083$ &
$9.876 \pm 0.351$ \\

MPC &
$\mathbf{0.033 \pm 0.001}$ &
$0.212 \pm 0.053$ &
$0.843 \pm 0.003$ &
$1.867 \pm 0.002$ \\

L1 &
$1.121 \pm 0.030$ &
$1.227 \pm 0.218$ &
$1.567 \pm 0.339$ &
$2.453 \pm 0.227$ \\

\textbf{Ours} &
$0.055 \pm 0.016$ &
$\mathbf{0.123 \pm 0.008}$ &
$\mathbf{0.433 \pm 0.010}$ &
$\mathbf{1.023 \pm 0.052}$ \\
\bottomrule
\end{tabular}
}
\endgroup
\end{table}

\begin{table}[h]
\centering
\caption{Trajectory-tracking RMSE for lawnmower trajectory in VRX simulator. The values in the first row refer to the set wind speed in the simulator.}
\label{tab:vrx_tracking_lawnmower}
\begingroup
\setlength{\tabcolsep}{2.5pt}
\renewcommand{\arraystretch}{0.9}
\resizebox{1.00\columnwidth}{!}{
\begin{tabular}{@{}l*{4}{c}@{}}
\toprule
Controller & 0.0 m/s & 1.0 m/s & 2.0 m/s & 3.0 m/s \\
\midrule

PID &
$1.327 \pm 0.025$ &
$1.453 \pm 0.036$ &
$1.983 \pm 0.036$ &
$2.708 \pm 0.022$ \\

MPC &
$0.022 \pm 0.001$ &
$0.022 \pm 0.001$ &
$0.053 \pm 0.022$ &
$0.131 \pm 0.000$ \\

L1 &
$0.079 \pm 0.015$ &
$0.108 \pm 0.052$ &
$0.075 \pm 0.033$ &
$0.075 \pm 0.014$ \\

\textbf{Ours} &
$\mathbf{0.016 \pm 0.002}$ &
$\mathbf{0.016 \pm 0.003}$ &
$\mathbf{0.029 \pm 0.001}$ &
$\mathbf{0.069 \pm 0.001}$ \\

\bottomrule
\end{tabular}
}
\endgroup
\end{table}

The PID controller exhibits the largest tracking errors overall. Although its integral action can reduce steady-state tracking bias, the controller relies only on tracking errors and does not explicitly incorporate the system dynamics. As a result, it cannot fully account for the vessel dynamics or disturbance effects, leading to degraded tracking performance as wind speed increases. This degradation is particularly evident for the zigzag trajectory, where the RMSE increases from 1.93 at 0 m/s to 9.88 at 3 m/s.

L1 adaptive MPC generally performs worse than nominal MPC, possibly due to over-estimation under relatively weak disturbances. However, in the lawnmower trajectory under 3 m/s wind, L1 outperforms nominal MPC, indicating improved disturbance compensation under stronger disturbances.

\subsection{Real-world Field Test}
\subsubsection{System Setup} 
\begin{figure}[t!]
    \centering
    \includegraphics[width=1.0\columnwidth]
    {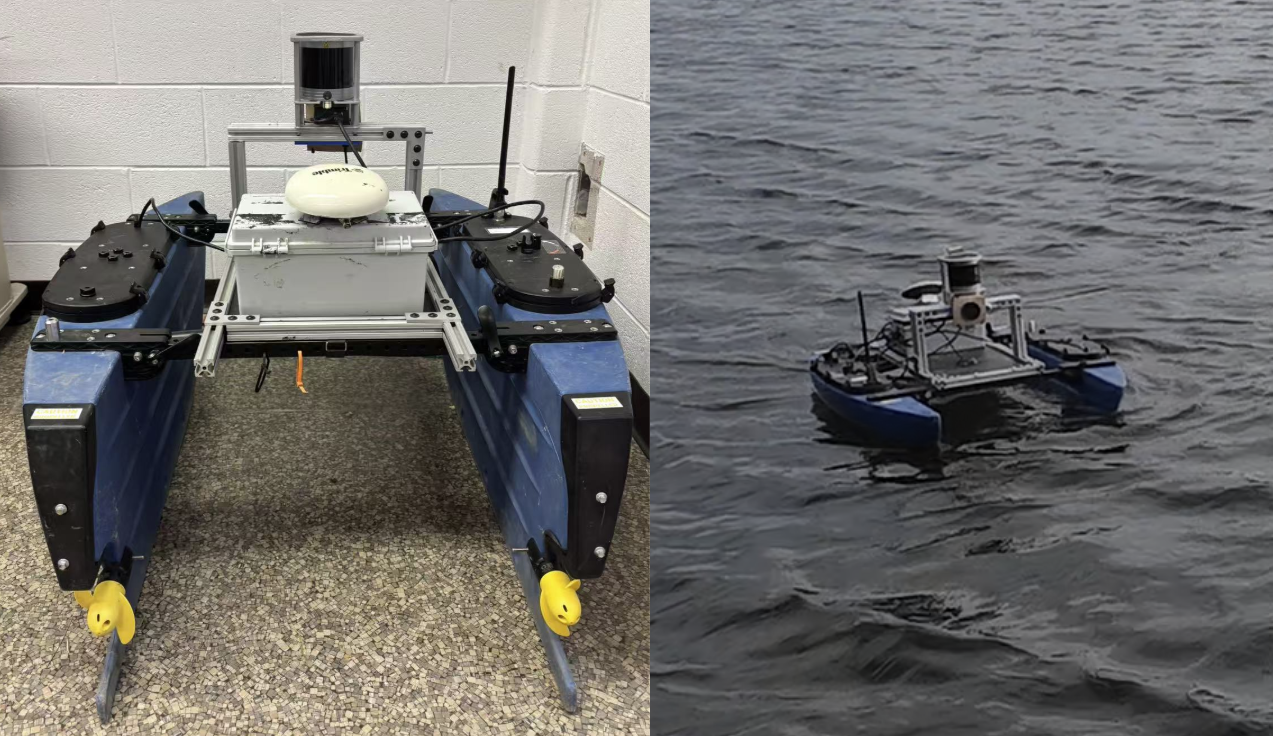}
    \caption{Real-world field test setup.}
    \label{fig:field_test_setup}
\end{figure}

As shown in Fig.~\ref{fig:field_test_setup}, the ASV used is a modified Blue Robotics BlueBoat. We leverage an Nvidia Jetson Orin AGX to handle on-board computation. The vessel, together with a sensor suite, has a mass of 28~kg and is actuated by two fixed parallel propellers with thrust limits of $[-26,50]$~N.

Field experiments were conducted in the same environment multiple times. We launch the boat in three different directions, and separately test the performance of the proposed method and baseline Lie-MPC. 
Although the river current is relatively consistent, the wind is time-varying in both magnitude and direction, leading to non-uniform disturbance conditions across trials. As a result, trajectories initialized from different directions (e.g., left-front vs. right-front) experience different effective disturbance profiles, which can lead to variations in tracking performance. In addition, minor drag from the safety rope and internal linearization errors also affect the field test outcomes.

\begin{figure}[t!]
    \centering
    \includegraphics[width=1.0\columnwidth]
    {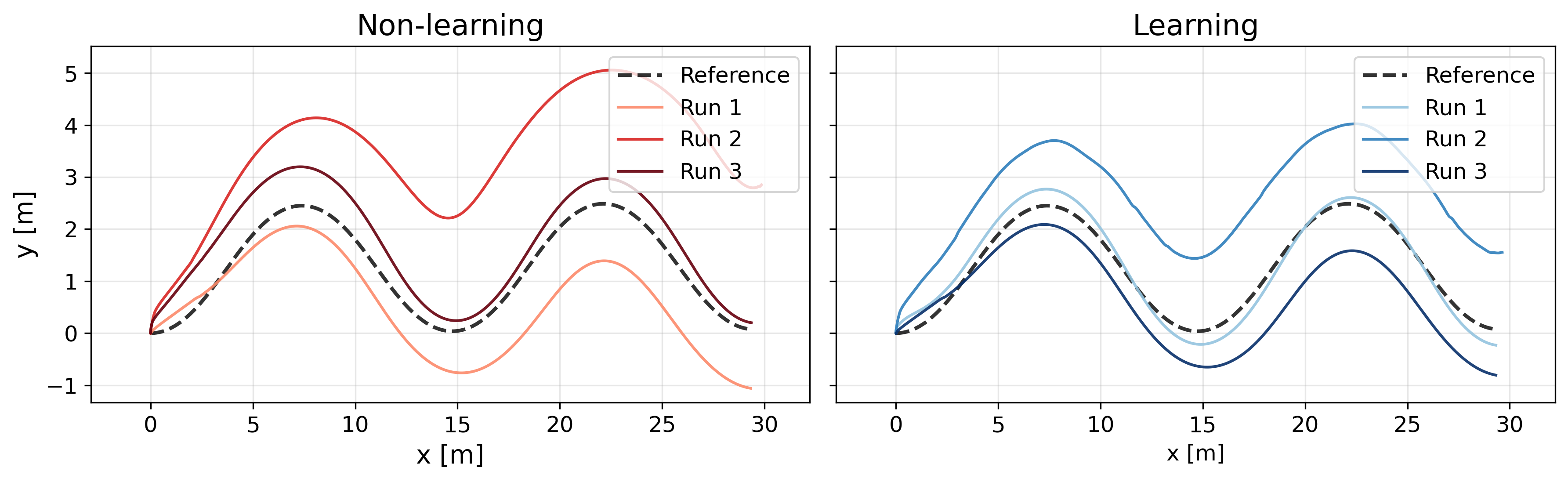}
    \caption{The trajectories in the field test.}
    \label{fig:real_trajectory}
\end{figure}

\subsubsection{Results}
As shown in Table~\ref{tab:vrx_tracking_summary}, Fig.~\ref{fig:real_trajectory}, the proposed Online-MPC outperforms the nominal MPC in zigzag trajectories. It achieves lower tracking errors and maintains stable operation under natural wind and current disturbances, demonstrating improved robustness and disturbance rejection capability in real-world conditions.

\begin{table}[!t]
\centering
\caption{Mean RMSE of trajectory tracking in field test.}
\label{tab:vrx_tracking_summary}
\footnotesize
\begin{tabular}{@{}lcc@{}}
\midrule
Controller & RMSE (m)  \\
\midrule
MPC     & $1.17 \pm 0.85$ \\
\textbf{Ours}        & {$\mathbf{0.76 \pm 0.56}$} \\
\midrule
\end{tabular}
\end{table}

\section{Conclusion}
We presented an Online Learning-based Model Predictive Control (Online-MPC) framework for trajectory tracking under unknown disturbances. By integrating online residual compensation into MPC, the controller compensates for unmodeled dynamics such as wind and hydrodynamic variations in real time.

Both simulation and field experiments demonstrate that Online-MPC outperforms the baselines, achieving lower tracking errors and greater robustness across diverse operating conditions. The results highlight the method’s effectiveness and practicality as a low-cost method for autonomous surface vessels operating in uncertain and dynamic environments.

Further study may focus on deploying this algorithm on other ASVs in an environment with obvious waves or evaluating the algorithm on a more task-based measurement for common ASV missions.

{
\balance
\bibliographystyle{IEEEtran}
\bibliography{bib/strings-abrv,bib/ieee-abrv,bib/references}

@article{zereik2018challenges,
  title={Challenges and future trends in marine robotics},
  author={Zereik, Enrica and Bibuli, Marco and Mi{\v{s}}kovi{\'c}, Nikola and Ridao, Pere and Pascoal, Ant{\'o}nio},
  journal={Annual Reviews in Control},
  volume={46},
  pages={350--368},
  year={2018},
  publisher={Elsevier}
}

@article{liu2016unmanned,
  title={Unmanned surface vehicles: An overview of developments and challenges},
  author={Liu, Zhixiang and Zhang, Youmin and Yu, Xiang and Yuan, Chi},
  journal={Annual Reviews in Control},
  volume={41},
  pages={71--93},
  year={2016},
  publisher={Elsevier}
}

@article{gao2024dynamic,
  title={Dynamic positioning control for marine crafts: a survey and recent advances},
  author={Gao, Xiaoyang and Li, Tieshan},
  journal={Journal of Marine Science and Engineering},
  volume={12},
  number={3},
  pages={362},
  year={2024},
  publisher={MDPI}
}

@article{veksler2016dynamic,
  title={Dynamic positioning with model predictive control},
  author={Veksler, Aleksander and Johansen, Tor Arne and Borrelli, Francesco and Realfsen, Bj{\o}rnar},
  journal={IEEE Transactions on Control Systems Technology},
  volume={24},
  number={4},
  pages={1340--1353},
  year={2016},
  publisher={IEEE}
}

@article{sarda2016station,
  title={Station-keeping control of an unmanned surface vehicle exposed to current and wind disturbances},
  author={Sarda, Edoardo I and Qu, Huajin and Bertaska, Ivan R and Von Ellenrieder, Karl D},
  journal={Ocean Engineering},
  volume={127},
  pages={305--324},
  year={2016},
  publisher={Elsevier}
}

@inproceedings{teng2022error,
  title={An error-state model predictive control on connected matrix {Lie} groups for legged robot control},
  author={Teng, Sangli and Chen, Dianhao and Clark, William and Ghaffari, Maani},
  booktitle=C-IROS,
  pages={8850--8857},
  year={2022},
  organization={IEEE}
}

@article{jang2023convex,
  title={Convex geometric trajectory tracking using {Lie} algebraic {MPC} for autonomous marine vehicles},
  author={Jang, Junwoo and Teng, Sangli and Ghaffari, Maani},
  journal={IEEE Robotics and Automation Letters},
  volume={8},
  number={12},
  pages={8374--8381},
  year={2023},
  publisher={IEEE}
}

@article{tu2019adaptive,
  title={Adaptive Control for Marine Vessels Against Harsh Environmental Variation},
  author={Tu, Fangwen and Ge, Shuzhi Sam and Choo, Yoo Sang and Hang, Chang Chieh},
  journal={arXiv preprint arXiv:1909.13265},
  year={2019}
}

@article{peng2023online,
  title={Online deep learning control of an autonomous surface vehicle using learned dynamics},
  author={Peng, Zhouhua and Xia, Fengbei and Liu, Lu and Wang, Dan and Li, Tieshan and Peng, Ming},
  journal={IEEE Transactions on Intelligent Vehicles},
  volume={9},
  number={2},
  pages={3283--3292},
  year={2023},
  publisher={IEEE}
}

@article{zhou2025simultaneous,
  title={Simultaneous system identification and model predictive control with no dynamic regret},
  author={Zhou, Hongyu and Tzoumas, Vasileios},
  journal={IEEE Transactions on Robotics},
  year={2025},
  publisher={IEEE}
}

@article{cho2020efficient,
  title={Efficient {COLREG}-compliant collision avoidance in multi-ship encounter situations},
  author={Cho, Yonghoon and Han, Jungwook and Kim, Jinwhan},
  journal={IEEE Transactions on Intelligent Transportation Systems},
  volume={23},
  number={3},
  pages={1899--1911},
  year={2020},
  publisher={IEEE}
}

@article{liu2020computationally,
  title={Computationally efficient {MPC} for path following of underactuated marine vessels using projection neural network},
  author={Liu, Cheng and Li, Cheng and Li, Wenhua},
  journal={Neural Computing and Applications},
  volume={32},
  number={11},
  pages={7455--7464},
  year={2020},
  publisher={Springer}
}

@article{annamalai2015robust,
  title={Robust adaptive control of an uninhabited surface vehicle},
  author={Annamalai, Andy SK and Sutton, Robert and Yang, Chenguang and Culverhouse, P and Sharma, S},
  journal={Journal of Intelligent \& Robotic Systems},
  volume={78},
  number={2},
  pages={319--338},
  year={2015},
  publisher={Springer}
}

@article{ghaffari2022progress,
  title={Progress in symmetry preserving robot perception and control through geometry and learning},
  author={Ghaffari, Maani and Zhang, Ray and Zhu, Minghan and Lin, Chien Erh and Lin, Tzu-Yuan and Teng, Sangli and Li, Tingjun and Liu, Tianyi and Song, Jingwei},
  journal={Frontiers in Robotics and AI},
  volume={9},
  pages={969380},
  year={2022},
  publisher={Frontiers Media SA}
}

@article{o2022neural,
  title={Neural-fly enables rapid learning for agile flight in strong winds},
  author={O’Connell, Michael and Shi, Guanya and Shi, Xichen and Azizzadenesheli, Kamyar and Anandkumar, Anima and Yue, Yisong and Chung, Soon-Jo},
  journal={Science Robotics},
  volume={7},
  number={66},
  pages={eabm6597},
  year={2022},
  publisher={American Association for the Advancement of Science}
}

@article{fossen1995nonlinear,
  title={Nonlinear modelling of marine vehicles in 6 degrees of freedom},
  author={Fossen, Thor I and Fjellstad, Ola-Erik},
  journal={Mathematical Modelling of Systems},
  volume={1},
  number={1},
  pages={17--27},
  year={1995},
  publisher={Taylor \& Francis}
}

@article{barrau2016invariant,
  title={The invariant extended {Kalman} filter as a stable observer},
  author={Barrau, Axel and Bonnabel, Silvere},
  journal={IEEE Transactions on Automatic Control},
  volume={62},
  number={4},
  pages={1797--1812},
  year={2016},
  publisher={IEEE}
}

@article{lale2024falcon,
  title={{FALCON: Fourier Adaptive Learning and Control for Disturbance Rejection Under Extreme Turbulence}},
  author={Lale, Sahin and Renn, Peter I and Azizzadenesheli, Kamyar and Hassibi, Babak and Gharib, Morteza and Anandkumar, Anima},
  journal={npj Robotics},
  volume={2},
  number={1},
  pages={6},
  year={2024},
  publisher={Nature Publishing Group UK London}
}

@incollection{hall2013lie,
  title={{Lie groups, Lie algebras, and representations}},
  author={Hall, Brian C},
  booktitle={Quantum Theory for Mathematicians},
  pages={333--366},
  year={2013},
  publisher={Springer}
}

@article{sola2018micro,
  title={A micro lie theory for state estimation in robotics},
  author={Sola, Joan and Deray, Jeremie and Atchuthan, Dinesh},
  journal={arXiv preprint arXiv:1812.01537},
  year={2018}
}

@book{bullo2019geometric,
  title={Geometric control of mechanical systems: modeling, analysis, and design for simple mechanical control systems},
  author={Bullo, Francesco and Lewis, Andrew D},
  volume={49},
  year={2019},
  publisher={Springer}
}

@article{singh2025gps,
  title={{GPS-DRIFT}: Marine Surface Robot Localization using {IMU-GPS} Fusion and Invariant Filtering},
  author={Singh, Surya Pratap and Lazouski, Tsimafei and Ghaffari, Maani},
  journal={arXiv preprint arXiv:2507.02198},
  year={2025}
}

@INPROCEEDINGS{1657243,
  author={Chengyu Cao and Hovakimyan, N.},
  booktitle=C-ACC, 
  title={Design and Analysis of a Novel {L1} Adaptive Controller, {Part I:} Control Signal and Asymptotic Stability}, 
  year={2006},
  volume={},
  number={},
  pages={3397-3402},
  doi={10.1109/ACC.2006.1657243}}

@article{teng2026lieipm,
  title={LieIPM: Lie Group Interior Point Method for Direct Trajectory Optimization of Rigid Bodies},
  author={Teng, Sangli and Zhang, Ruiqi and Lin, Tzu-Yuan and Clark, William A and Mueller, Mark and Vasudevan, Ram and Ghaffari, Maani and Sreenath, Koushil},
  journal={arXiv preprint arXiv:2606.10579},
  year={2026}
}

@STRING{C-ACC = {Proc. Amer. Control Conf.}}

@STRING{C-IROS = {Proc. {IEEE}/{RSJ} Int. Conf. Intell. Robots and Syst.}}
}

\end{document}